%% file: main.tex
\documentclass[fleqn,10pt]{wlscirep}
\usepackage[utf8]{inputenc}
\usepackage[T1]{fontenc}
\usepackage{mathtools}
\usepackage{times}
\usepackage{latexsym}
\usepackage{multicol}
\usepackage{graphicx}
\usepackage{algorithm}
\usepackage[noend]{algpseudocode}
\usepackage{varwidth}
\usepackage{multirow}
\usepackage{hyperref}
\usepackage{makecell} 
\usepackage{makecell} 
\usepackage{amsfonts}

\input{pref}

\title{Machine Learning Optimal Ordering in Global Routing Problems in Semiconductors}

\author[1,+]{Heejin Choi}
\author[1,+]{Minji Lee}
\author[1]{Chang Hyeong Lee}
\author[2]{Jaeho Yang}
\author[1,*]{Rak-Kyeong Seong}

\affil[1]{Department of Mathematical Sciences, Ulsan National Institute of Science and Technology, Ulsan, South Korea}
\affil[2]{Samsung SDS,  AI Advanced Research Lab,
  Samsung R\&D Campus, Seocho-Gu, Seoul, South Korea}
  
\affil[*]{seong@unist.ac.kr}

\affil[+]{these authors contributed equally to this work}

\begin{document}

\begin{abstract}
In this work, we propose a new method for ordering nets during the process of layer assignment in global routing problems. 
The global routing problems that we focus on in this work are based on routing problems that occur in the design of substrates in multilayered semiconductor packages. 
The proposed new method is based on machine learning techniques and we show that the proposed method supersedes conventional net ordering techniques based on heuristic score functions. 
We perform global routing experiments in multilayered semiconductor package environments in order to illustrate that the routing order based on our new proposed technique outperforms previous methods based on heuristics. 
Our approach of using machine learning for global routing targets specifically the net ordering step which we show in this work can be significantly improved by deep learning. 
\end{abstract}  
\maketitle


\section{Introduction}

Routing in semiconductor chip and package design is the process of creating and designing electrical connections between different components of a chip or package, such as transistors, gates, input and output pins, solder balls and other components depending on the design specifications of the chip and package. 
For example, for a Fine Pitch Ball Grid Array (FBGA) package with up to approximately 500 pins, an expert human designer may take up to 48 hours to complete a routing design for the semiconductor package. This does not take into account design testing for the design candidate which might end up sub-optimal for manufacturing. 
As a result, design automation and optimization in semiconductor chip and package design has been playing a significant role for the last few decades \cite{gao2008new, lee2008congestion,chang2010nthu, lee2009robust, kahng2022machine, chan2017routability} in order to simplify the task of routing design in semiconductor chip and packages. 
This field of research is also known today as electronic design automation (EDA) \cite{lim2008practical,wang2009electronic,lavagno2016electronic} in computer-aided design (CAD). 

Automating and optimizing the routing process is a quintessential problem in EDA for semiconductor chips and packages.
This is because the design complexity that results from a given routing solution directly impacts the performance, power consumption, signal integrity, manufacturability and costs of producing a semiconductor chip and package \cite{lavagno2016electronic,lavagno2018eda}.
Moreover, with an increasing number of components to be connected by non-intersecting electrical connections in a 3-dimensional chip and package environment, the routing problem becomes exponentially more difficult not just to automate, but also to optimize against design and performance constraints. 

With this exponentially increasing difficulty in mind, a series of innovations \cite{10.1145/3451179} in EDA introduced the concept of global routing \cite{albrecht2001global,wu2010grip, behjat2006integer,chen2009high,cho2009boxrouter,moffitt2008maizerouter} with the aim of simplifying and isolating the problem of routing from the rest of the design problem for semiconductor chips and packages.
The main idea is to separate the routing problem into two parts: \textit{global routing} and \textit{detailed routing} \cite{yutsis2014ispd,zhou2015accurate,qi2014accurate,spindler2007fast}.
In global routing, the locations of areas that need to be connected during the routing process are embedded into a grid graph which discretizes and simplifies the multi-layered routing environment. After connections are obtained in global routing, the grid is mapped back during detailed routing to the original routing environment by implementing design constraints such as wire-widths and wire separations. 

In this work, we concentrate on the process of global routing in multi-layered routing problems, especially in semiconductor package substrate design. 
In particular, we focus on the layer assignment problem in global routing \cite{hsu2008multi, lee2008congestion, wu2004layer,ciesielski1989layer}. 
The layer assignment problem occurs when a multilayered routing problem is simplified as a routing problem in a 2-dimensional grid. When the routing solution is found in the 2-dimensional grid, the routing solution is mapped to a $k$-layered grid graph by assigning connections between layers and layer numbers to the original 2-dimensional routing solution.
The corresponding algorithm was first introduced in \cite{lee2008congestion}.
There it was noted that if the routing problem consisted of several nets, the layer assignment of the 2-dimensional routing solutions for each individual net heavily depended on the order in which each net solution was assigned layers during the layer assignment process. 

A solution to the ordering optimization problem proposed in the original work in \cite{lee2008congestion} was to assign each net and its 2-dimensional routing solution a heuristic score according to which a net order can be computed. 
This heuristic score function depended on features such as the total routing length, overflow and vertices used in the 2-dimensional routing solution.
In this work, we propose a new method of ordering nets and their 2-dimensional routing solution for the layer assignment process in global routing. Our method is based on machine learning a collection of features of the 2-dimensional routing solutions and to predict the most optimal net ordering that results in the most desired 3-dimensional routing solution after layer assignment. 
Our proposed new method supersedes and outperforms the heuristics-based net ordering method in \cite{lee2008congestion} based on experimental results that we present in the following work. 
\\

\section{Problem Formulation \label{sec:2}}

\begin{figure}[ht!]
\begin{center}
\resizebox{0.9\hsize}{!}{
\includegraphics[height=6cm]{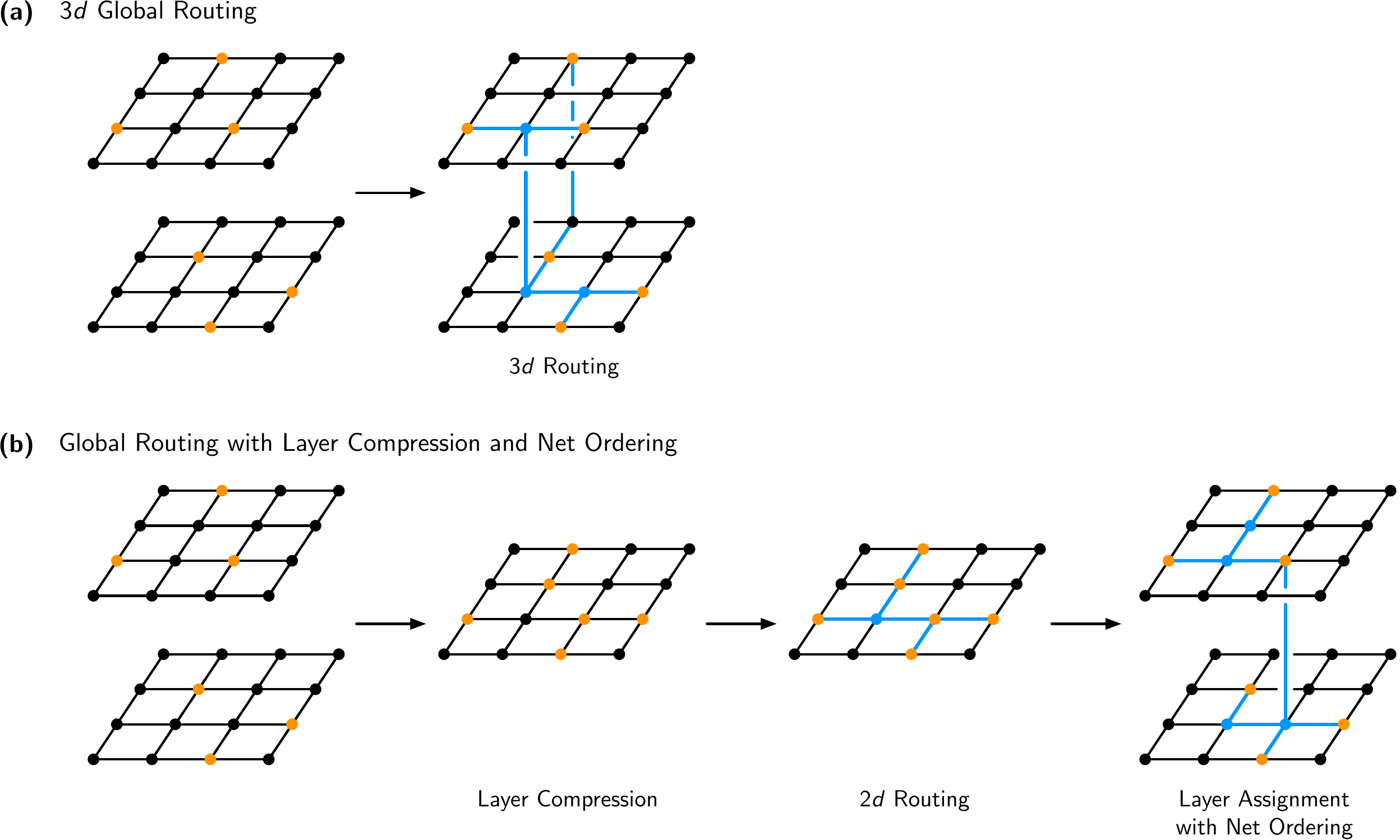} 
}
\caption{
Comparison between (a) 3-dimensional global routing directly inside the multilayered environment, and (b) global routing with layer compression and net ordering.
\label{fcomparison}}
 \end{center}
 \end{figure}

Multilayered routing is a critical problem in very large scale integration (VLSI) technology in semiconductor chip and package design \cite{lim2008practical,wang2009electronic,lavagno2016electronic}.
Here, routing is conducted in two phases known as global routing \cite{hsu2008multi, lee2008congestion, wu2004layer,ciesielski1989layer}
and detailed routing \cite{sechen1985timberwolf,shin1986mighty}.
In the following work, we concentrate on global routing which can be conducted in two different approaches that are both illustrated in \fref{fcomparison}.
The first approach is to directly find the routing in the multilayered routing environment, where the objects to be connected are distributed in different layers effectively requiring a routing in 3-dimensional space.
Even though this approach outputs directly a 3-dimensional multilayered routing result with the benefit that the cost of connecting different layers can be directly taken into account in the 3-dimensional routing process, 
this approach is computational expensive becoming exponentially harder to solve with more layers and more nets in the problem.
The second approach is to compress the multilayered routing environment into a single layer grid on which one can solve the routing problem first and then assign to each section of the routing a layer in order to get the 3-dimensional multilayered routing solution. 
Here, the problem becomes much more computational solvable, however it introduces a new decision step known as net ordering which is the main topic of discussion in this work.

\begin{figure}[ht!]
\begin{center}
\resizebox{0.8\hsize}{!}{
\includegraphics[height=6cm]{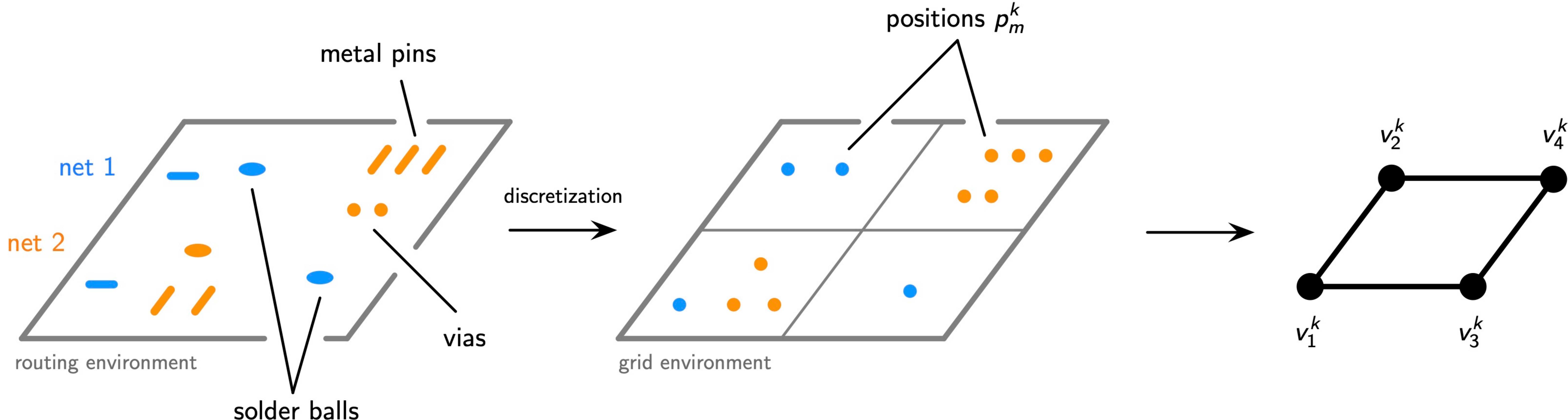} 
}
\caption{
Process of Discretization during Global Routing, where physical objects in the routing environment are interpreted as vertices in a grid graph $G^k$.
\label{fig03}}
 \end{center}
 \end{figure}

\begin{figure}[ht!]
\begin{center}
\resizebox{0.9\hsize}{!}{
\includegraphics[height=6cm]{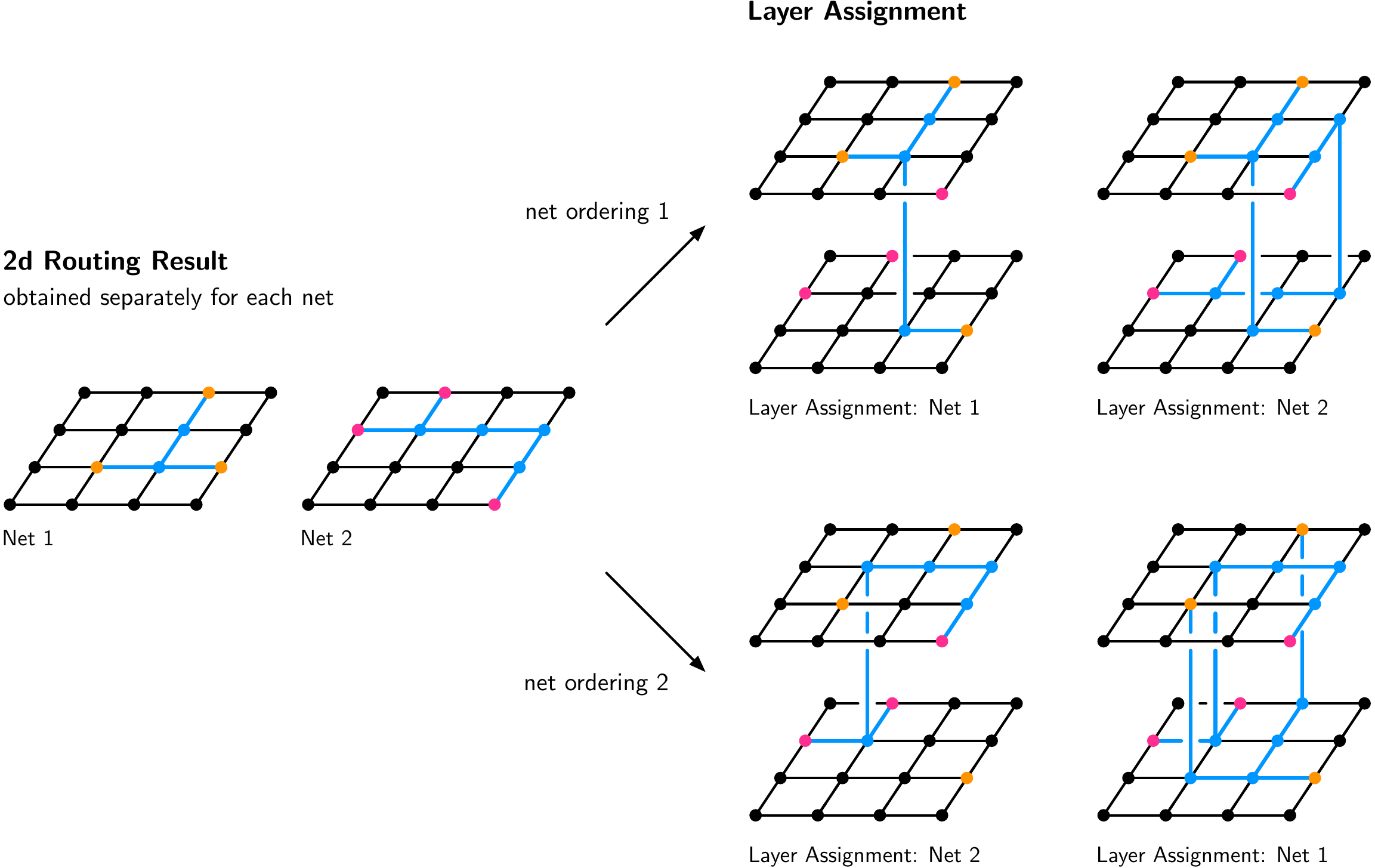} 
}
\caption{
Net ordering during layer assignment of the $2d$ routing results affects the overall $3d$ routing result in global routing. 
\label{fnetordering}}
 \end{center}
 \end{figure}

In the following, we give a brief summary of the global routing process using the second approach with layer compression and layer assignment, and give a brief overview of the optimization problem involved in the net ordering.

\begin{figure}[ht!]
\begin{center}
\resizebox{0.65\hsize}{!}{
\includegraphics[height=6cm]{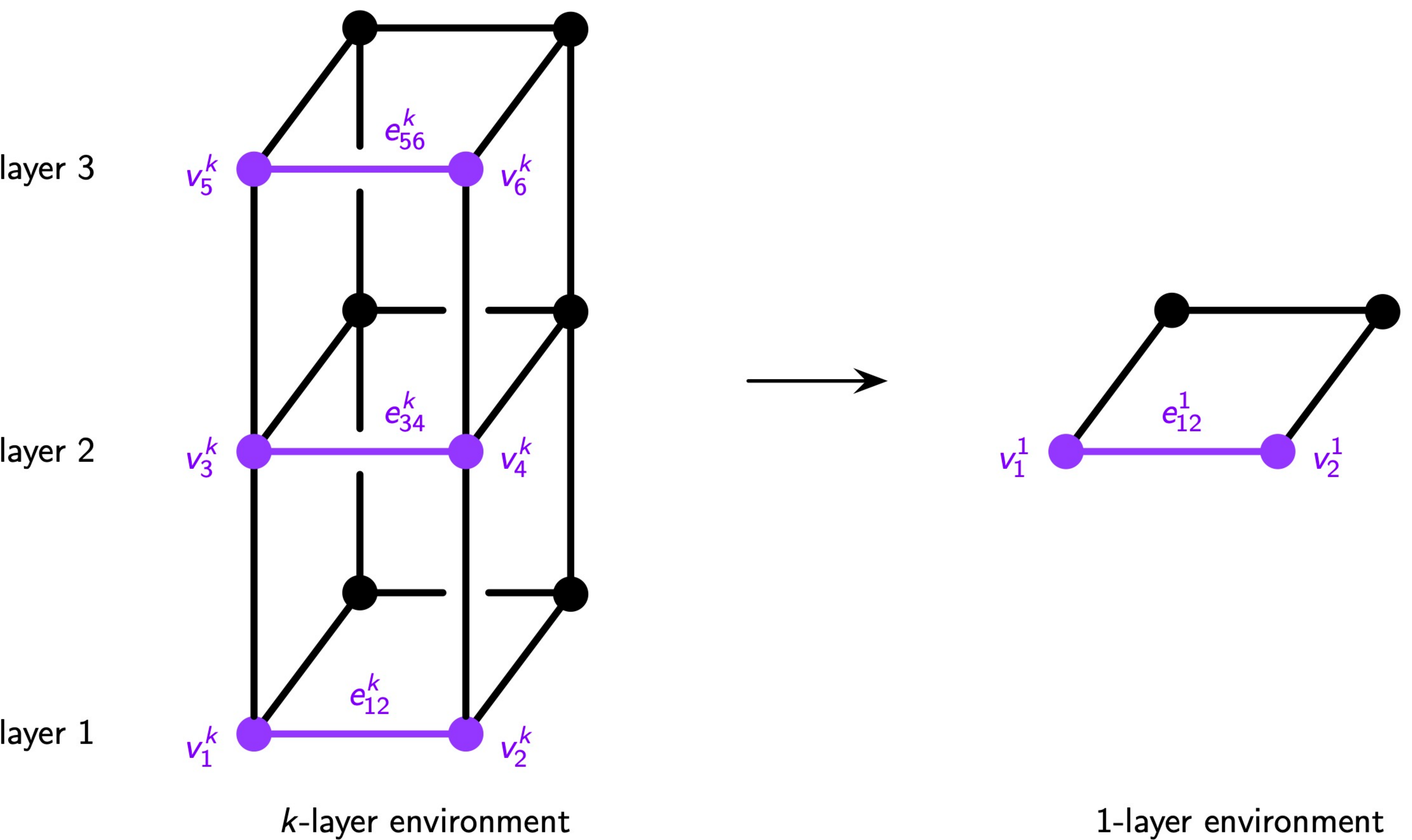} 
}
\caption{
$1$-layer compression for a $3$-layer grid graph. Under the compression, equivalent vertices $v_m^k$ and edges $e_{ij}^k$ are mapped to single vertices $v_m^1$ and edges $e_{ij}^1$ in the 1-layer compressed graph $G^1$.
\label{fig02}}
 \end{center}
 \end{figure}

\paragraph{Global Routing Algorithm based on Layer Compression.}
Global routing in a $k$-layer routing environment involves the following algorithmic steps:
\begin{itemize}
\item[1.] \underline{Grid Graph:} We obtain the $k$-layer grid graph $G^k = (V^k,E^k)$ from the 3-dimensional routing environment and routing problem.
The grid vertices $v_i \in V^k$ correspond to grid regions in the $k$-th layer of the routing environment. 
Physical objects such as solder balls and metal pins on the routing environment that need to be connected are represented by pins $p_m^k$, which are each associated to one of the grid vertices $v_i \in V^k$.
The boundary separating two neighboring grid regions is mapped to an edge $e_{ij}^k = (v_i^k, v_j^k)$ in $G^k$.
Next to \textit{boundary edges} $E_b^k$ that connect neighboring grid vertices on the same layer, we also have \textit{via edges} $E_v^k$ that connect neighboring grid vertices in adjacent layers. 
An edge $e_{ij}^{k}$  is assigned a \textit{capacity} $c(e_{ij}^k) \in \mathbb{Z}^+$, which is the number of available connections that are possible, and the \textit{length} $l(e_{ij}^k) \in \mathbb{R}^+$, which the physical wirelength in the actual routing environment for any connection that passes through $e_{ij}^k$. In this work, we will assume that all edge wirelengths are the same such that for all $e_{ij}^k, e_{uv}^k \in E^k$, $l(e_{ij}^k) = l(e_{uv}^k)$. \fref{fig03} shows an example of a routing environment divided into $2\times 2$ grid regions that correspond to vertices $v_1^k, \dots, v_4^k$ in a grid graph.
\item[2.] \underline{Routing Problem and Nets:} Vertices $v_i \in V^k$ belong to nets $n_m^k \in N^k$, and vertices corresponding to the same net have to be connected during the routing process.  Given a net $n_m^k \in N^k$, a \textit{global routing solution} on layer $k$ of the routing environment is a connected subgraph $T_m^k = (V_m^k, E_m^k)$ of the grid graph $G^k$, where
$V_m^k \subset V^k$ and $E_m^k \subset E^k $ are respectively the subset of vertices and edges that contribute to the global routing solution for $n_m^k$.
We note that for a given net $n_m^k$, there can be more than one routing solution $T_m^k$. 
If every net $n_m^k$ in the global routing problem is assigned a routing solution $T_m^k$, the collection $S^k = \{T_m^k\}$ is known as the \textit{complete global routing solution} for $(G^k, N^k)$. 

\item[3.] \underline{Layer Compression:} 
1-layer compression takes the $k$-layer grid graph $G^k = (V^k,E^k)$ and maps it to $ G^1=(V^1,E^1)$. \fref{fig02} illustrates the $1$-layer compression for a 3-layer grid graph. This reduction into a 2-dimensional routing problem still encapsulated the constraints given in the 3-dimensional routing problem.
These preserved constraints include the \textit{compressed capacity}, which is $c(e_i^1) = \sum_{e_m^k \in ES(e_i^1)} c(e_m^k)$. 

\item[4.] \underline{$2d$ Global Routing:} Solve the $2$-dimensional routing problem by finding the $1$-layer routing solution $S^1 = \{T_m^1\}$. Solutions are found using algorithms such as the \textit{Kruskal's Algorithm (KA)} \cite{kruskal1956shortest}, based on the search algorithm for the \textit{minimum spanning tree} along the grid graph $G^1$, and the 
\textit{Steiner Tree Algorithm (ST)} \cite{kou1981fast} based on an algorithm for identifying the shortest possible tree formed by a subset of edges in $G^1$.

\item[5.] \underline{Net Ordering:} Every net in the 1-layer compression is connected by the $2d$ routing result. In order to now assign layers to each routing segment in $T_m^1$, we have to first identify in which \textit{order} nets are assigned layer information. Depending on the net ordering, the $3$-dimensional global routing result will become different. This process is known as \textit{net ordering} and is illustrated in \fref{fnetordering}.
\item[6.] \underline{Layer Assignment:} The $2$-dimensional routing solution $S^1 = \{T_m^1\}$ is lifted to a solution on the $k$-layer grid  $S^k=\{T^k_m\}$ by a process called \textit{layer assignment}.
\end{itemize}
In the following work, we will put a particular emphasis on the last two steps of the global routing process known as \textit{net ordering} and \textit{layer assignment}.
\\

\paragraph{Net Ordering.}
The order in which subgraphs $T_m^1$ for each net $n_m^1$ are lifted to $3$ dimensions affects dramatically the overall $3d$ global routing solution on $G^k$. 
This is because given a certain finite non-zero capacity $c(e_{ij}^k) \in \mathbb{Z}^+$ on all edges $e_{ij}^k \in E^k$ in the $3$-dimensional grid graph $G^k$, the $3d$ global routing solution $T_m^k$ after layer assignment should not contain a grid graph $G^k$ with edges that have a \textit{overflow} given by,
\beal{es13a01}
o(e_{ij}^k ) = \left\{ 
\ba{cc}
d(e_{ij}^k) - c(e_{ij}^k) & \text{if $d(e_{ij}^k)> c(e_{ij}^k)$}
\\
0 & \text{otherwise}
\ea
\right.
~,~
\eea
where $d(e_{ij}^k)$ is the number of subgraphs $T_m^k$ containing edge $e_{ij}^k$, i.e. the \textit{demand} for edge $e_{ij}^k$, and $c(e_{ij}^k)$ is the capacity of edge $e_{ij}^k$.
Depending on which subgraph $T^1_m$ is lifted first to $T^k_m$, the distribution of demand in the $k$-layer graph $G^k$ changes significantly. 
With the design constraint that overflow is kept at $o(e_{ij}^k )=0$ for all edges $e_{ij}^k \in E^k$, the distribution of demand $d(e_{ij}^k)$ determined by the ordering of $T_m^1$ heavily affects the overall $k$-layer global routing result $S^k$. 

As a result, net ordering significantly affects the overall 3-dimensional global routing solution $S^k$. We define a \textit{net ordering} as a ordered sequence,
\beal{es13a10}
(T_{m_1}^1 , T_{m_2}^1, \dots , T^1_{m_{N_{\text{nets}}}}) ~:~
\text{score}(T_{m_1}^1) \geq \text{score}(T_{m_2}^1) \geq \dots \geq \text{score}(T_{m_{N_{\text{nets}}}}^1) ~,~
\eea
where $N_{\text{nets}}$ is the number of nets in $N^1$, and $\text{score}(T_{m}^1) \in \mathbb{R}_{\geq 0}$ is a score value assigned to each 1-layer routing solution $T_m^1$ corresponding to net $n_m^1$, which is used for the net ordering.
In \cite{lee2008congestion}, the score function was defined heuristically for $T_m^1=(V_m^1,E_m^1)$ as follows
\beal{es13a11}
\text{score}(T_m^1)
= \frac{\alpha}{l(T_m^1)} + \beta N_{\text{pins}}(T_m^1) + \gamma \overline{\rho}(T_m^1) ~,~
\eea
where $l(T_m^1)$ is the total wirelength for the subgraph $T_m^1$ defined by $l(T_m^1) = \sum_{e_{ij}^1 \in E_m^1} l(e_{ij}^1)$.
$N_{\text{pins}}(T_m^1) = |P_m^1|$ is the number of pins in $P_m^1$ of the subgraph $T_m^1$ corresponding to net $n_m^1$.
Furthermore, the last term in \eref{es13a11} depends on the average net density $\rho(T_m^1)$ for the subgraph $T_m^1$, which is defined as follows
\beal{es13a13}
\overline{\rho}(T_m^1) 
= \frac{
\sum_{e_{ij}^1 \in E_m^1} d(e_{ij}^1)
}{
\sum_{e_{ij}^1 \in E_m^1} c(e_{ij}^1)
}
~,~
\eea
where $d(e_{ij}^1)$ is the demand on the edge $e_{ij}^1$ in the 1-layer routing solution $S^1$ and $c(e_{ij}^1)$ is the capacity of the edge $e_{ij}^1$.
According to \cite{lee2008congestion}, the coefficients $\alpha$, $\beta$ and $\gamma$ are determined by the designed constraints of the $3d$ routing problem.

We will see in the following work that a net ordering algorithm based on machine learning outperforms the heuristic function in \eref{es13a11} in various experiments. 
\\

\paragraph{Layer Assignment.}
We review here briefly an algorithm known as \textit{single-net optimal layer assignment (SOLA)} based on \cite{lee2008congestion}.
We start with an $k$-layer grid $G^k$ where all edges of the $k$-layer grid have demand $d(e_{ij}^k) = 0$ and overflow $o(e_{ij}^k)=0$.
Through layer assignment of a compressed $2d$ global routing solution $S^1 = \{ T_m^1 \}$ with net ordering $(T_{m_1}^1 , T_{m_2}^1, \dots , T^1_{m_{N_{\text{nets}}}})$, 
we go sequentially through the subgraphs $T_{m}^1$ and layer assign their edges into $G^k$. This process updates $d(e_{ij}^k) $ with the constraint that $o(e_{ij}^k)=0$ has to be kept for all edges. 

\begin{figure}[ht!]
\begin{center}
\resizebox{0.65\hsize}{!}{
\includegraphics[height=6cm]{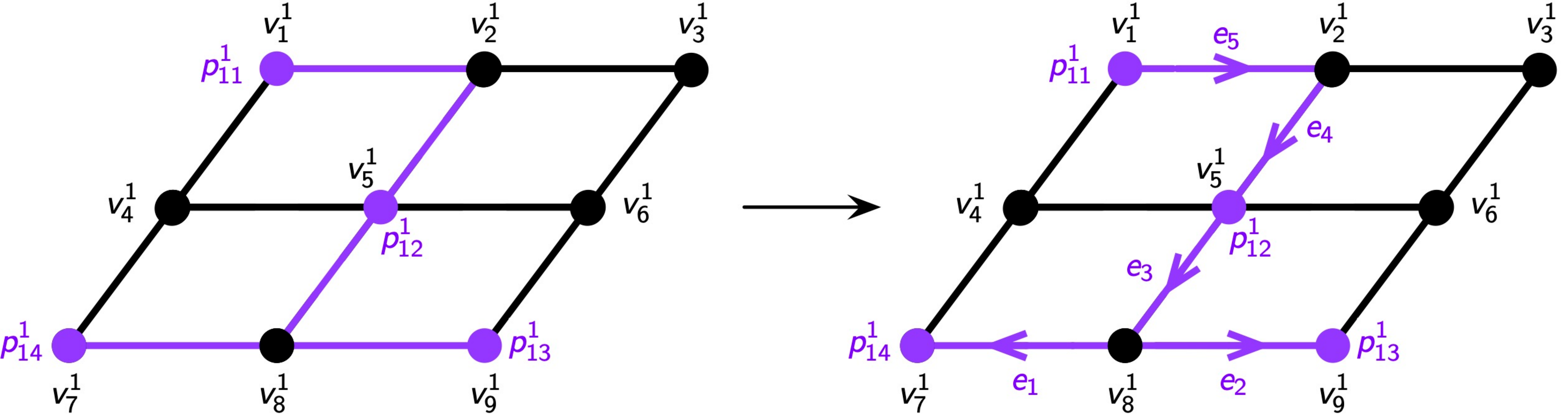} 
}
\caption{
An orientation is assigned to edges $e_{ij}$ of a minimum spanning tree $T_m^1$ by choosing $p_{11}^1$ as the source and all other vertices as sinks of the orientation. 
\label{fig07}}
 \end{center}
 \end{figure}

Given the minimum spanning tree $T_{m}^1 = (V_m^1, E_m^1)$, the edges in $E_m^1$ can be assigned directions such that one of the pins $p_{ij}^\text{source}$ in $P_m^1$ is the source where all edges begin from, and all other pins $p_{ij}^\text{sink}$ in $P_m^1$ are the sinks where all edges end. 
This orientation of edges in $ E_i^1$ is illustrated in a simple example in \fref{fig07}.
Under such an orientation of edges, every vertex in $v_i^1 \in V_m^1$ has a natural set of children vertices defined as 
\beal{es13a20}
ch(v_i^1) = \{ 
w ~|~ \vec{e} = (v_i^1, w) ~,~ \vec{e} \in E_m^1
\}
~,~
\eea
where $\vec{e}$ is an oriented edge starting from vertex $v_i^1$ and ending at vertex $w$. 
We can also define the parent vertex for $v_i^1$ as 
\beal{es13a21}
par(v_i^1) = w ~,~ \vec{e} = (w,v_i^1) \in E_m^1 ~,~
\eea
where $\vec{e}$ is an oriented edge starting from vertex $w$ and ending at vertex $v_i^1$. 
In this case, we call $\vec{e} = (w,v_i^1)$ also the parent edge $par_e(v_i^1)$

We now take the $1$-layered $2d$ routing solution to be positioned at the $k=1$ layer of the $k$-layer grid graph $G^k$.
Using the orientation of edges for a given subtree $T_m^1$, we can now define a recursive heuristic cost function which measures the cost of moving a parent edge $par_e(v_i^1)$ for a vertex $v_i^1$ to layer $r$ of the $k$-layered graph $G^k$,
\beal{es13a22}
mvc(v_i^1, r ) 
= \min_{
\ba{c}
1\leq r_1 \leq k \\
\vdots \\
1\leq r_q \leq k 
\ea
}
\left(
\sum_{j=1}^q 
~mvc(v_j^1, r_j) + vc(v_i^1)
\right) ~,~
\eea
where $q=|ch(v_i^1)|$ is the number of children vertices of $v_i^1$ and $vc(v_i^1)$ is the cost of adding a via edge between the first layer and the $r$-th layer. 
We note that the sum in \eref{es13a22} is over all children vertices of $v_i^1$ labelled by $j=1,\dots, |ch(v_i^1)|$.
The key of the SOLA algorithm is to minimize the overall cost $mvc(v_i^1,r)$ for all vertices $v_i^1$ in subtree solution $T_m^1$, with the constraint that all edge overflows are kept at zero, $o(e_{ij}^k) =0$.

\begin{figure}[ht!]
\begin{center}
\resizebox{0.2\hsize}{!}{
\includegraphics[height=6cm]{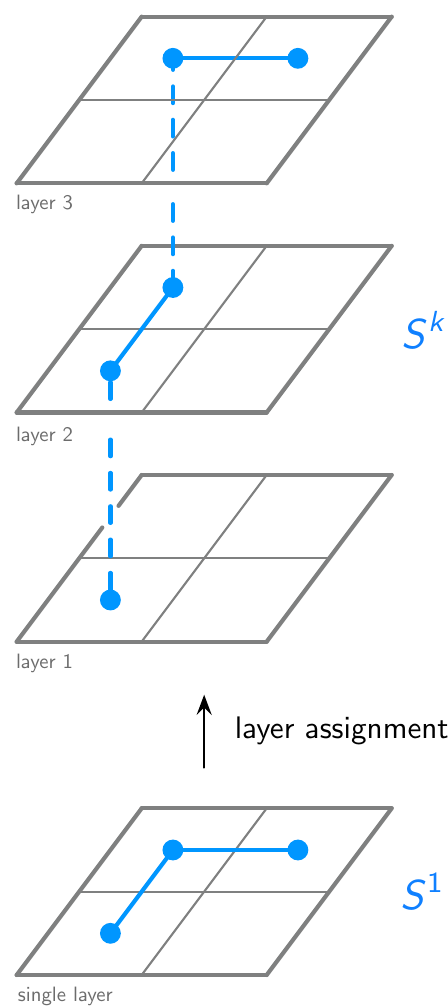} 
}
\caption{
Layer assignment of a 1-layer routing solution $S^1$ to a $k$-layer routing solution $S^k$.
\label{fig08}}
 \end{center}
 \end{figure}

Using this recursive score function, every edge in a subgraph $T_m^1$ can be lifted to an optimal layer $r$.
The addition of via edges that connect vertices between different layers, the SOLA algorithm ensures that the resulting $T_m^k$ subgraph in $G^k$ is a connected between layers. 
\fref{fig08} illustrates an example where a 1-layer routing solution $S^1$ is lifted to a $k$-layer routing solution $S^k$.

We make use of the SOLA algorithm to obtain for a given specific net ordering a $k$-layer routing solution $S^k= \{T_m^k \}$. 
In addition to SOLA, there is a complementary algorithm known as \textit{Accurate and Predictable Examination for Congestion} (APEC) \cite{lee2008congestion}, which minimizes the overall overflow in the routing solution during layer assignment. Combined with SOLA, the two algorithms are known as \textit{Congestion-Constrained Layer Assignment} (COLA) \cite{lee2008congestion}, which we use in our experiments in this work.
By comparing different routing solutions $S^k(T_{m_1}^1 , T_{m_2}^1, \dots , T^1_{m_{N_{\text{nets}}}})$ for different net orderings $(T_{m_1}^1 , T_{m_2}^1, \dots , T^1_{m_{N_{\text{nets}}}})$, we can identify which net ordering results in the most optimal $k$-layer global routing solution.
In the following sections, we present how we trained a deep neural network to predict the most optimal net ordering $(T_{m_1}^1 , T_{m_2}^1, \dots , T^1_{m_{N_{\text{nets}}}})$ that results in the most optimal $k$-layer global routing solution $S^k$. 
We show that our machine learning model outperforms the heuristics-based net ordering scoring function in \eref{es13a11} by conducting several experiments. 
\\

\section{Method \label{sec:3}}

\paragraph{Features for Machine Learning. \label{sec:features}}
In this work, we present a machine learning model that takes as an input the collection of $1$-layer $2d$ global routing solutions $S^1 = \{T_m^1\}$ and outputs the most optimal net ordering $(T_{m_1}^1 , T_{m_2}^1, \dots , T^1_{m_{N_{\text{nets}}}})$ which results in the most optimal $k$-layer routing solution $S^k = \{T_m^k\}$.
In order to train the deep neural network, we identify the following features from the $1$-layer global routing solutions $T^1_m = (V_m^1, E_m^1)$:
\begin{itemize}
	
\item \underline{Number of Pins $|P^k_m|$: } This is the number of pins associated to vertices $v_i^k \in V^k$ and a given net $n^k_m$ in a $k$-layer routing environment. 

\item \underline{Number of Pins $|P^1_m|$: } This is the number of pins associated to compressed vertices $v_i^1 \in V^1$ and a given net $n^1_m$ in a single layer routing environment prior to layer assignment. 

\item \underline{Number of Vertices $|V^1_m|$: } This is the number of vertices in $V^1_m$ in the $2d$ routing solution $T^1_m$ for a given net $n^1_m$.

\item \underline{Overflow $o(E_m^1)$: } This is the total overflow of the edges $e_{ij}^m$ in $E_m^1$ in the $2d$ routing result $T^1_m$ for a given net $n^1_m$. The total overflow is given by
$o(E_m^1) = \sum_{e_{ij}^m \in E_m^1} o(e_{ij}^k)$,
where the overflow $o(e_{ij}^k)$ for an edge $e_{ij}^k$ is defined in \eref{es13a01}.

\item \underline{Minimum Rectangle: } 
We define a minimum rectangle that contains all of the vertices in the $2$-dimensional routing solution $S^1$.
This rectangle has the following dimensions
\beal{es20a00b}
&
\Delta x_{min} =  \max_{v_i^1 \in V^1} x(v_i^1) - \min_{v_j^1 \in V^1} x(v_j^1) ~~,~~
\Delta y_{min} =  \max_{v_i^1 \in V^1} y(v_i^1) - \min_{v_j^1 \in V^1} y(v_j^1) ~,~
&
\eea
where $x(v_i^1)$ and $y(v_i^1)$ are the $x$- and $y$-coordinates of the vertex $v_i^1 \in V^1$, respectively. 
The area of the rectangle is given by $A_{min} = \Delta x_{min} \cdot \Delta y_{min}$.

\item \underline{Number of Branch Vertices $|V_{\text{branch}}^1|$: } 
This is the number of vertices in the 2-dimensional routing solution $S^1$, which are branch points in $T^1$. 
A vertex $v_i^1 \in V^1$ is a branch point if $|ch(v_i^1)|>1$ or $|par(v_i^1)|>1$.

\end{itemize}

In summary, for a given $k$-layer routing solution $S^k$, we can obtain a feature vector of the following form,
\beal{es20a00d}
&&
\vec{f}(S^k) = 
\left(
\{|P_m^k|\},
\{|P_m^1|\},
\{|V_m^1|\},
\{o(E_m^1)\},
\Delta x_{min},
\Delta y_{min},
A_{min},
|V^1_{\text{branch}}|
\right)
~,~
\eea
where the index $m=1,\dots, N_{\text{nets}}$ labels the nets in $S^k$.
\\

\paragraph{Optimal Net Order. \label{sec:score}}
Our goal is to find the optimal net order based on criteria used in substrate routing to evaluate optimality in routing designs
 \cite{ISPD2008}. 
 The criteria used in \cite{ISPD2008} to evaluate optimality uses features of the routing results $S^k$ after layer assignment.

 In our work, we simplify the optimality criteria in \cite{ISPD2008}.
 Given a $2d$ routing solution $S^1$, we use a candidate net ordering $no=(T_{m_1}^1 , T_{m_2}^1, \dots , T^1_{m_{N_{\text{nets}}}})$ to obtain the corresponding layer assigned $k$-layer routing solution $S^k$ in order to evaluate the optimality of the $k$-layer routing solution $S^k$ as follows:
\begin{itemize}
	\item[1.] Given two $k$-layer routing solutions $S^k_1$ and $S^k_2$ corresponding to net orderings $no_1$ and $no_2$ associated to the same $2d$ routing solution $S^1$, we calculate the total overflow for $S^k_1$ and $S^k_2$, which is given by
	$o(S^k) = \sum_{e_{ij}^k \in E^k} o(e_{ij}^k)$,
	where $E^k$ is the set of edges in the layer assigned routing solution $S^k$, and $o(e_{ij}^k)$ is the overflow for edge $e_{ij} \in E^k$, which is defined in \eref{es13a01}.
	The most optimal $k$-layer routing solution is determined by the one which has the smaller value for $o(S^k)$. If $o(S^k_1)=o(S^k_2)$, we calculate the maximum overflow
	for a routing solution instead. The maximum overflow $mo(S^k)$ for a routing solution $S^k$ is defined as 
	$mo(S^k) = \max_{e_{ij}^k \in E^k} o(e_{ij}^k)$.
	If for two routing solutions $S^k_1$ and $S^k_2$ we have $o(S^k_1)=o(S^k_2)$, then the most optimal $k$-layer routing solution $S^k$ is determined by the one which has the smaller value for $mo(S^k)$. 
	
	\item[2.] If for two routing solutions $S^k_1$ and $S^k_2$ we have both $o(S^k_1)=o(S^k_2)$ and $mo(S^k_1)=mo(S^k_2)$, then we calculate the optimality score value which is defined as follows, 
	\beal{es20a01}
	\text{Score}(S^k) = l(S^k) \times (1+ t_{run})~,~
	\eea 
	where $L$ is the total wirelength of edges in $S^k$, 
	$l(S^k) = \sum_{e_{ij}^k \in E^k} l(e_{ij}^k)$, 
	and $t_{run}$ is the processor runtime for layer assignment of $S^1$ to $S^k$.
	A smaller $\text{Score}(S^k)$ corresponds to a more optimal $k$-layer routing solution $S^k$. 
	All experiments in this work are conducted on a server with Intel Xeon CPU E5-2687W. 
\end{itemize}
Based on the above measures, we can assign an optimality value for each $k$-layer routing solution $S^k$ that was obtained from a net ordering $no(S^k)$ for the layer assignment of $S^1$. 
For a single layer routing solution $S^1$ with $N_{\text{nets}}$ number of nets, the total number of candidate net orderings is given by 
$N_{\text{ordering}} = (N_{\text{nets}})!$.
The optimality of $S^k$ takes an integer value in $[1,N_{\text{ordering}}]$ which represents the rank from most optimal, \textit{i.e.} $\text{optimality}(S^k)=1$, to least optimal, \textit{i.e.} $\text{optimality}(S^k)=N_{\text{ordering}}$. 
The actual optimality value and ranking between two $k$-layer routing solutions $S^k_1$ and $S^k_2$ is decided by the following conditions based on the optimality measures discussed above,
\beal{es20a06}
\text{optimality}(S^k_1) < \text{optimality}(S^k_2) \Leftrightarrow 
\left\{
\ba{lc}
1. & o(S^k_1) \leq o(S^k_2)\\
2. &mo(S^k_1) \leq mo(S^k_2)\\
3. &\text{Score}(S^k_1) < \text{Score}(S^k_2)
\ea
\right.
~.~
\eea

\paragraph{Machine Learning Model. \label{sec:model}}
In \cite{lee2008congestion}, optimal net ordering was determined by a heuristic score function which we reviewed in \eref{es13a11}.
This static function assigns a score for each net $n_m^1$ in 2-dimensional routing solution $S^1$ and the net ordering is determined by the ordering of the score values. 
In our work, we replace the heuristic method in \cite{lee2008congestion} with a deep learning model that orders the nets in $S^1$ in order to obtain the most optimal $k$-layer routing solution after layer assignment. 
Because our deep learning model is trained to result in the most optimal $k$-layer routing solution after layer assignment, it is qualitatively different to the heuristic ordering method in \cite{lee2008congestion}, which only depends on static features of the 1-layer routing solution $S^1$ without receiving any feedback from the $k$-layer routing solution $S^k$ after layer assignment. 

In order to train our deep-learning model, 
we generate for a given routing problem a 1-layer routing solution $S^1$ and construct from it the $k$-layer routing solutions $S^k_h$ by using all possible net orderings $no_{h}$, where $h=1, \dots, N_{\text{ordering}}$ labels the individual net orderings.
For each of these $k$-layer routing solutions $S^k_h$ obtained by layer assignment of $S^1$ with a given net ordering $no_h$, we measure the optimality of $S^k_h$ using the process explained in Section \sref{sec:score}. 
Furthermore, we extract the feature vector $\vec{f}(S^k_h)$ summarized in Section \sref{sec:features} of the $k$-layer routing solutions $S^k$. 
Both the features vector $\vec{f}(S^k_h)$ and $\text{optimality}(S^k_h)$ are used to train our deep learning model.

The training data is segmented into groups. 
Each training data group corresponds to a given 1-layer routing solution $S^1$ and consists of all possible net orders labelled by $h=1, \dots, N_{\text{ordering}}$.
By computing the $k$-layer routing solution $S^k_h$ for each of the net orderings $no_h$, 
the net orders $no_h$ and the corresponding solution $S^k_h$ are ordered by determining $\text{optimality}(S^k_h)$.
We also calculate for each solution $S^k_h$ the features vector $\vec{f}(S^k_h)$.
Our deep learning model takes for each $S^k_h$ per data group as input the feature vector $\vec{f}(S^k_h)$ and outputs the optimality rank $\text{optimality}(S^k_h)$ for the net ordering $no_h$ within the data group corresponding to $S^1$. 

Although our training data is labelled by $\text{optimality}(S^k_h)$, the ranking done using $\text{optimality}(S^k_h)$ is within each data group corresponding to a single layer routing solution $S^1$, so that only the nets and net ordering $no_h$ within the same group affect the ranking under $\text{optimality}(S^k_h)$. 
This means that even with the same feature values from both the $1$-layer routing solution $S^1$ and $k$-layer routing solution $S^k$, the best net ordering under $\text{optimality}(S^k_h)$ can differ depending on the different net orderings $no_h$ within the group corresponding to $S^1$. 
In order to address this issue, our deep learning model consists of two parts: 
\begin{itemize}
\item[(1)] 
\underline{Learning between Data Groups:}
The aim here is to train the deep learning model by using feature vectors $\vec{f}(S^k_h)$ across different data groups associated to different single layer routing solutions $S^1$.
\item[(2)] 
\underline{Learning within a Data Group:}
The aim here is to train the deep learning model by using feature vectors $\vec{f}(S^k_h)$ within a given data group associated to a single routing solution $S^1$. 
\end{itemize}

Our deep learning model uses a multilayer perceptron (MLP) architecture. 
This is a fundamental and widely used deep learning model that makes use of a variety of activation and loss functions. 
Since the output of the last layer is the optimal net ordering $no$ based on the ranking by $\text{optimality}(S^k_h)$,
our deep learning model can be identified as a classification model. 
Accordingly, we use Mean Squared Error (MSE) and Cross-Entropy (CE), 
which are often used in classification models as loss functions. 
The MSE is defined as follows,
\beal{es21a01}
\text{Mean Squared Error (MSE)} = \frac{1}{N}\sum_{i=1}^{N}(y_i-\hat{y_i})^2
~,~
\eea
where $N$ is the size of the labelled dataset, $y_i$ is the expected output and $\hat{y_i}$ is the predicted output for the $i$-th datapoint.
Similarly, the Cross-Entropy is defined as 
\beal{es21a02}
\text{Cross-Entropy (CE)} = -\frac{1}{N}\sum_{i=1}^{N}\sum_{j=1}^{C} \sum_{m=1}^{C} p_{ijm} \log(q_{ijm}) ~,~
\eea
where in our deep learning models $N$ is the number of data groups each corresponding to a single layer solution $S^1_i$, 
$C$ corresponds to the number of nets $N_\text{nets}$,
and $p_{ij}$ is the probability $N_\text{nets}$-vector
that 
indicates the actual optimal order $m$ of the net $n^k_j$ with a probability component $p_{ijm}=1$, whereas all other components for the net $n^k_j$ are taken to be $0$.
In comparison, $q_{ij}$ is the predicted probability $N_\text{nets}$-vector which labels with a component $q_{ijm}$ the predicted probability that the net $n^k_j$ has order $m$.
Note that the $\log$-function in \eref{es21a02} is used to penalize more heavily incorrect probability predictions than correct ones. 

\begin{figure}[H]
\begin{center}
\resizebox{0.4\hsize}{!}{
\includegraphics[height=6cm]{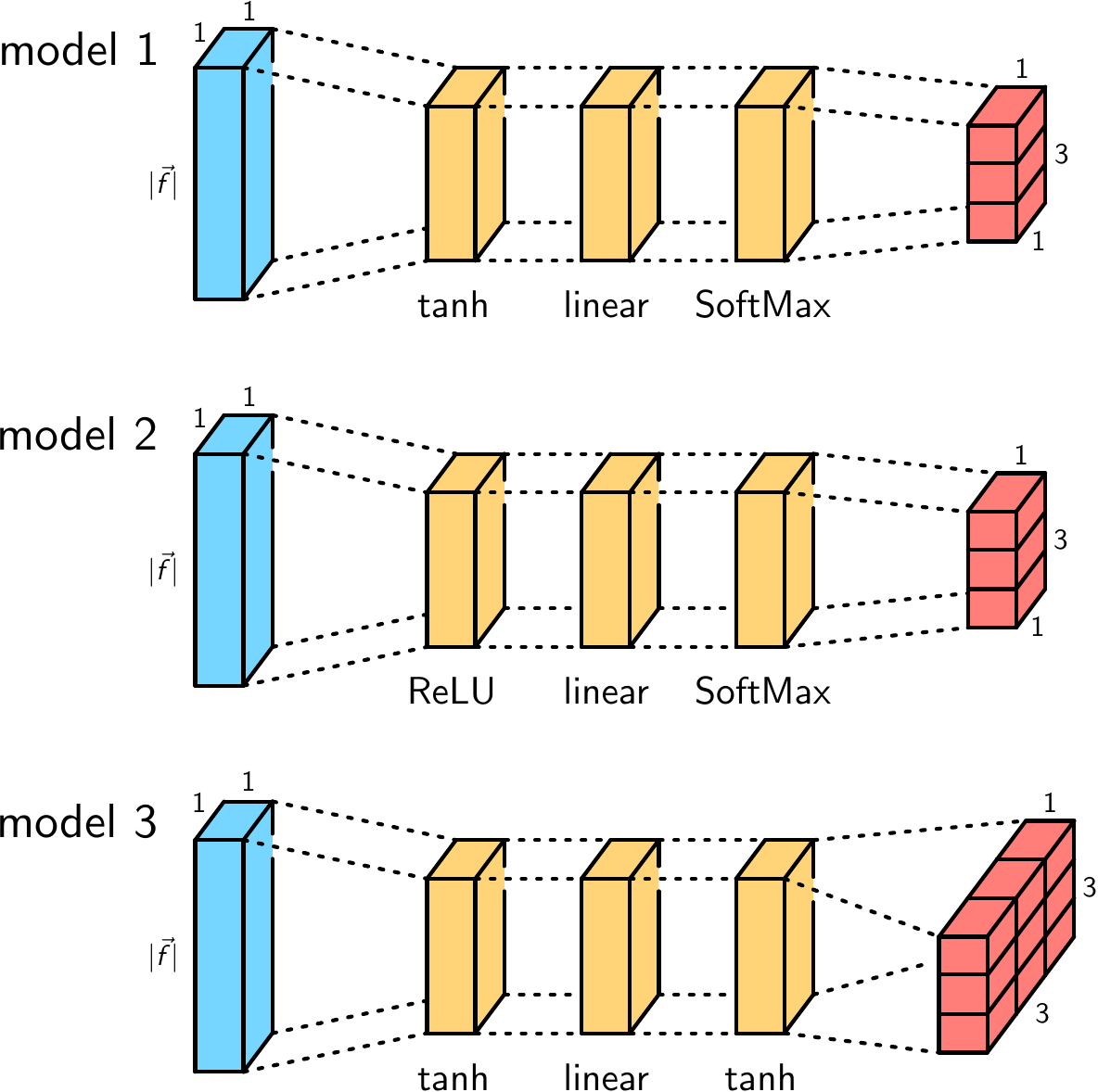} 
}
\caption{
Architectures of the 3 deep learning model used to predict the most optimal net ordering. 
\label{fig:fmodel}}
 \end{center}
 \end{figure}

In total, we propose 3 deep learning models for the task of identifying the optimal net ordering for a single layer routing solution $S^1$.
These 3 deep learning models with their layer configurations and loss functions are summarized in \tref{tab:t03} and are illustrated in \fref{fig:fmodel}.
We selected a basic deep learning architecture, a multilayer perceptron (MLP), to ensure accessibility and simplicity. 
The MLP is a fundamental neural network model that is widely understood and can be easily adapted for various use cases. 
In order to predict the net ordering value through deep learning, we utilized two types of output layers: a softmax activation function for Models 1 and 2, and a hyperbolic tangent (tanh) function for Model 3.
Model 1 and Model 2 return values for $\text{optimality}(S^k_h)$ for each net ordering $no_h$ in a data group corresponding to a solution $S^1$. 
Based on the value for $\text{optimality}(S^k_h)$, the optimal net ordering is determined within a data group associated to $S^1$. 
Here, the MSE defined in \eref{es21a01} is used to minimize the difference between the actual best net ordering with the predicted net ordering during the model training process. 
In comparison, Model 3 returns a probability value for each net ordering $no_h$ in a data group corresponding to $S^1$.
The probability is a measure of whether the predicted $no_h$ is the actual optimal net ordering for $S^1$. 
Using the cross-entropy defined in \eref{es21a02}, the model is trained to maximize the probability $q_{ij}$ of the predicted optimal net ordering $no_j$ across different data groups corresponding to different single layer solutions $S^1_i$. 
Furthermore, in order to explore the impact of different activation functions in the first layer, Model 1 was chosen to use the ReLU function and Model 2 was chosen to use the tanh function for comparison.

\begin{table}[ht!]
\begin{center}
\resizebox{0.7\hsize}{!}{
\begin{tabular}{|c|c|c|c|c|}
\hline
Model & \makecell{Layer 1 \\ (Activation function)} & Layer 2 & \makecell{Layer 3 \\ (Activation function)} & Loss function\\
\hline\hline
Model 1 & \makecell{Linear \\ (Hyperbolic Tangent)} & \makecell{Linear} & \makecell{Linear \\ (Softmax)} & MSE \\
\hline
Model 2 & \makecell{Linear \\ (ReLU)} & \makecell{Linear} & \makecell{Linear \\ (Softmax)} & MSE \\
\hline
Model 3 & \makecell{Linear \\ (Hyperbolic Tangent)} & \makecell{Linear} & \makecell{Linear \\ (Hyperbolic Tangent)} &  Cross-Entropy \\
\hline 
\end{tabular}	
}
\caption{
Summary of the 3 different types of deep learning models used to predict the most optimal net ordering.
\label{tab:t03}
}
\end{center}
\end{table}

\section{Experiment and Data \label{sec:4}}

\subsection{Data Preparation \label{sec:41}}

\begin{table}[ht!]
\begin{center}
\begin{tabular}{|c|c|}
\hline
dataset parameter & options\\
\hline\hline
$1$-layer router type & Kruskal algorithm(KA), Steiner tree(ST)\\
number of layers $k$ & 2, 5\\
number of nets $N_{\text{nets}}$ & 3, 5\\
feature vector & all features ($\vec{f}$), reduced features($\vec{f}_\text{reduced}$)\\
\hline 
\end{tabular}
\caption{
Summary of parameters that define the routing problem conditions for each dataset during dataset generation.
\label{tab:t04}
}
\end{center}
\end{table}

\begin{figure}[ht!]
	\begin{center}
		\resizebox{0.85\hsize}{!}{
			\includegraphics[trim={0cm 0cm 0cm 0cm}, width=0.8 \linewidth]{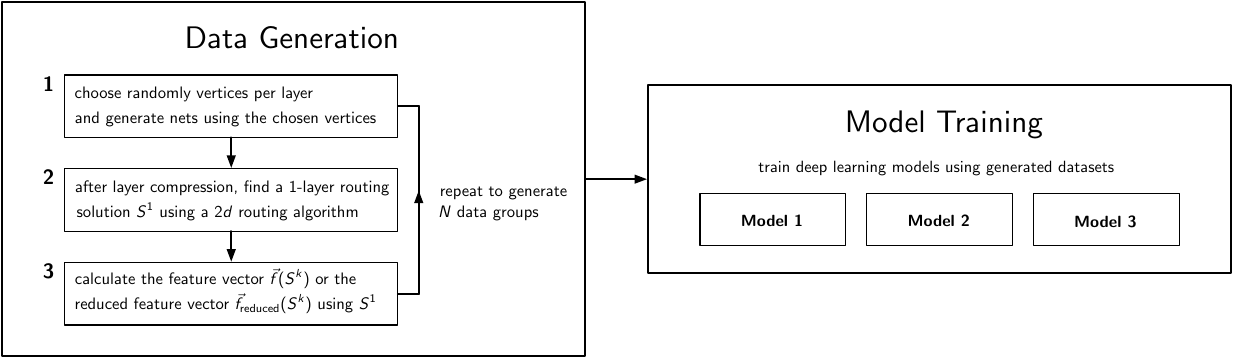}
		}
		\caption{
			The data preparation process and the training process for the deep learning model.
			\label{fig:datapre}
		}
	\end{center}
\end{figure}
	
\fref{fig:datapre}
briefly shows the process of generating datasets and how to use the generated datasets as input data in our deep learning model. In the dataset preparation part, we computationally generate the $16$ independent training datasets before training the deep learning model. One dataset includes $N = 2500$ data groups, where each data group corresponds to a different $1$-layer routing solution $S^1_i$ for a corresponding randomly generated routing problem. 
For a given data group, we first randomly choose a constant number of vertices per layer and assign the selected vertices to the number of nets $N_{\text{nets}}$ at random. Secondly, we find the 1-layer routing solution $S^1$ through layer compression using 1-layer routing algorithms such as KA, and ST. 
Thirdly, the feature vector $\vec{f}(S^k)$ or reduced feature vector $\vec{f}_{\text{reduced}}(S^k)$ of the 1-layer 2$d$ routing solution $S^1$ is calculated based on the definition of the feature vector discussed in Section {\sref{sec:features}}. 
Finally, after repeating the above steps for $N$ times, a single dataset is generated. 
We then move on to training our 3 deep learning summarized in \tref{tab:t04} and illustrated in \fref{fig:fmodel} using the feature vector $\vec{f}(S^k)$ or reduced feature vector $\vec{f}_{\text{reduced}}(S^k)$ generated previously as part of the dataset. 

For training, we use in total $16$ independent generated datasets that are determined by $4$ conditions, which are varied as discussed as follows:
\begin{itemize}
\item \underline{Number of layers $k$:} We vary the total number of layers $k$ in the routing environment. The $16$ datasets are generated such that for $8$ datasets we have $k=2$ layers in the routing environment and for the other $8$ datasets we have $k=5$ layers in the routing environment.

\item \underline{Number of nets $N_{\text{nets}}$:} We vary the total number of nets $N_{\text{nets}}$ in the $k$-layer routing problem in order to generate the $16$ independent datasets. $8$ of the datasets are obtained by setting $N_{\text{nets}}=3$, while the other remaining $8$ datasets are obtained by setting $N_{\text{nets}}=5$.

\item \underline{Single Layer Router:}
Before layer assignment, we have to first have a 1-layer routing solution $S^1$ with a choice of net ordering $no$.
The 1-layer routing solution $S^1$ is found by one of two algorithms, the first being the Kruskal's algorithm (KA) and the second being the Steiner Tree algorithm (ST).
$8$ of the datasets have been generated using Kruskal's algorithm and the other $8$ datasets have been generated using the Steiner Tree algorithm for $S^1$.  

\item \underline{Selection of routing features:}
As we discussed in Section \sref{sec:features}, we can assign a feature vector $\vec{f}(S^k_h)$ for a given $k$-layer routing solution $S^k_h$ obtained from a net ordering $no_h$ as defined in \eref{es20a00d}.
This feature vector contains as components the number of pins $|n^k_m|$ in the $k$-layer environment, the number of pins $|n^1_m|$ in the single layer environment, the number of vertices $|V^1_m|$ in the single layer environment, the overflow $o(E_m^1)$ in the single layer environment, the number of branch vertices $|V^1_{\text{branch}}|$ in the single layer environment, and information on the minimal encompassing rectangle as defined in \eref{es20a00b} for the single layer environment.
We note that most of the features in $\vec{f}(S^k_h)$ depend on the single layer routing solution $S^1$ rather than information relating to the $k$-layer routing problem.

As argued in \cite{jurado2015hybrid, zheng2017training, hwang2020using}, 
a proper optimal selection of features often leads to a better training result for deep learning models. 
Accordingly, when we generate the training datasets, we also allow an option for a narrower selection of features.
Being always motivated to choose features from the single layer routing solution $S^1$ that highly correlate with features of the $k$-layer routing solution $S^k_h$ for a given net ordering $no_h$, 
we select a subset of features \cite{jebli2021prediction, luong2022feature} discussed in Section \sref{sec:features} that correspond to the $1$-layer solution $S^1$.
This subset forms a reduced feature vector $\vec{f}_{\text{reduced}}(S^k_h)$ for a given $k$-layer routing solution $S^k_h$, which we define as follows
\beal{es22a01}
\vec{f}_{\text{reduced}}(S^k_h)=
\left(
\{ |n^k_m| \}
,
\{ |n^1_m| \}
,
\{ |V^1_m| \}
,
\{ o(E^1_m) \}
\right)
~,~
\eea
where $m=1, \dots, N_{\text{pins}}$ is the index over nets $n_m^1$ in the single layer environment. 
For $8$ datasets, we calculate $\vec{f}(S^k_h)$ as the feature vector, and for the remaining $8$ dataset we use the reduced feature vector $\vec{f}_{\text{reduced}}(S^k_h)$. 

\end{itemize}
\tref{tab:t04} summarizes the above parameters.

\begin{figure}[ht!]
	\begin{center}
		\resizebox{0.75\hsize}{!}{
			\includegraphics[trim={0cm 0cm 0cm 0cm}, width=0.8 \linewidth]{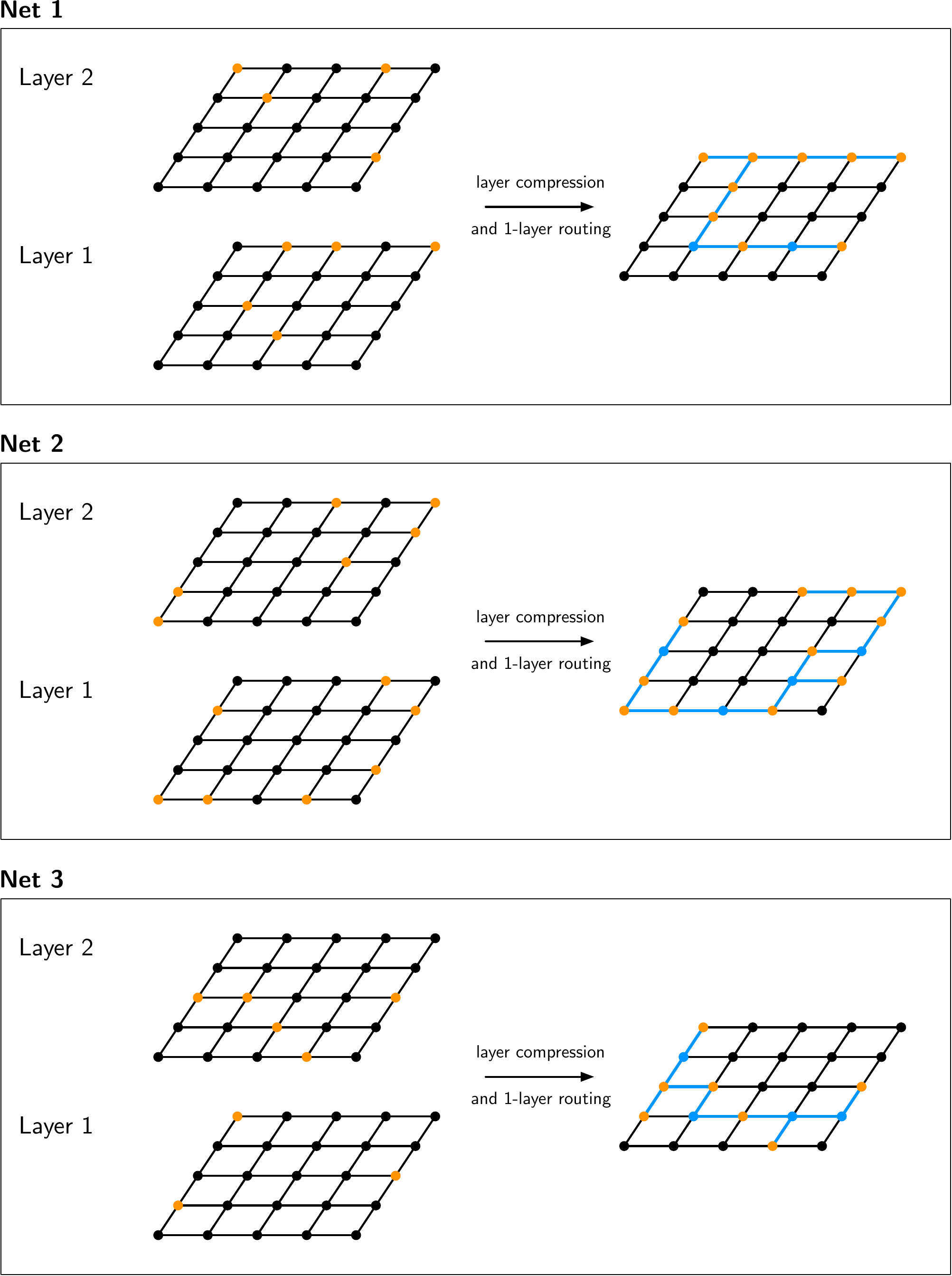}
		}
		\caption{
			1-layer routing solution $S^1$ for a $k=2$ layer routing environment with $N_{\text{nets}}=3$. The vertices $v_m^k$ corresponding to pins are highlighted in orange and are restricted to a constant number of 15 per layer. The 1-layer routing solution $S^1$ was obtained through the Steiner tree algorithm.}
			\label{fig:ddataex}
	\end{center}
\end{figure}

A complete summary of the $16$ datasets generated by varying the parameters described above is given in \tref{tab:t05}. 
Given a dataset generated for a combination of fixed conditions, as described above, 
we generate $N=2500$ data groups each corresponding to a 1-layer routing solution for randomly generated routing problems. 
The routing problems are restricted to a $k$-layer routing environment where each layer consists of a $5\times 5$ grid graph $G^1$. 
Furthermore, the vertices $v_m^k$ that correspond to pins in the routing problem and are part of the connected graph in the routing solution are restricted such that the number of them is a constant 15 per layer.
This ensures that in the generated datasets are made of $k$-layer routing solutions $S^k$ that evenly involve 15 connected vertices $v_m^k$ per layer. The 1-layer routing solution $S^1$ of one sample data group in Data 3 is shown in \fref{fig:ddataex}.
\\

\begin{table}[ht!]
\begin{center}
\begin{tabular}{|c|c|c|c|c|}
\hline
dataset & 1-layer router type & $k$-layers & $N_{\text{nets}}$ & features\\
\hline\hline	
Data 1 & & & & all \\		
Data 2 & \multirow{-2}*{KA} & \multirow{-2}*{2} & \multirow{-2}*{3} & reduced\\
\hline 		
Data 3 & & & & all\\		
Data 4 & \multirow{-2}*{ST} & \multirow{-2}*{2} & \multirow{-2}*{3} & reduced\\
\hline 		
Data 5 & & & & all\\		
Data 6 & \multirow{-2}*{KA} & \multirow{-2}*{5} & \multirow{-2}*{3} & reduced\\
\hline 		
Data 7 & & & & all\\		
Data 8 & \multirow{-2}*{ST} & \multirow{-2}*{5} & \multirow{-2}*{3} & reduced\\
\hline 		
Data 9 & & & & all\\		
Data 10 & \multirow{-2}*{KA} & \multirow{-2}*{2} & \multirow{-2}*{5} & reduced\\
\hline 		
Data 11 & & & & all\\		
Data 12 & \multirow{-2}*{ST} & \multirow{-2}*{2} & \multirow{-2}*{5} & reduced\\
\hline 		
Data 13 & & & & all\\		
Data 14 & \multirow{-2}*{KA} & \multirow{-2}*{5} & \multirow{-2}*{5} & reduced\\
\hline 		
Data 15 & & & & all\\		
Data 16 & \multirow{-2}*{ST} & \multirow{-2}*{5} & \multirow{-2}*{5} & reduced\\
\hline 
\end{tabular}
\caption{
Summary of generated dataset with choice of parameters that are used to generate the sample routing solutions and their features. 
\label{tab:t05}
}
\end{center}
\end{table}

\subsection{Experiments and Model Training \label{sec:42}}

We divide the generated dataset into a $80\%$ training set and a $20\%$ testing set. 
The test data is used to evaluate the accuracy of our proposed deep learning models. 
Our deep learning models were trained by using the Adam optimizer and by varying various hyperparameter values of the models. 
For hyperparameter tuning, a grid search was used to train our models while changing each hyperparameter value by a certain amount. 
This approach allows us to identify optimal configurations that enhance model performance and generalizability. 
By exploring the hyperparameter space, we aimed to mitigate the risk of overfitting, ensuring that our models not only perform well on the training data but also exhibit robust performance on unseen data.
\tref{tab:t06} summarized the values of each hyperparameter that was used for hyperparameter tuning during model training.

\begin{table}[ht!]
\begin{center}
\resizebox{0.7\hsize}{!}{
\begin{tabular}{|c|c|}
\hline
hyperparameter & values \\
\hline\hline
number of epochs & 30, 50, 70, 90, 100, 150, 200, 500, 1000, 1500, 2000\\
number of units & 10, 20, 30 ,40, 50, 60, 70, 80, 100\\
learning rate & 0.0001, 0.0005, 0.001, 0.002, 0.003, 0.004, 0.005, 0.008, 0.01 \\
\hline
\end{tabular}
}
\caption{
Hyperparameter values used during hyperparameter tuning of the 3 deep learning models.
The unit number corresponds to the layer width of the hidden layers in the neural network as discussed in Section \sref{sec:model}.
\label{tab:t06}
}
\end{center}
\end{table}

\begin{figure}[ht!]
\begin{center}
\resizebox{0.75\hsize}{!}{
\includegraphics[trim={0cm 0cm 0cm 0cm}, width=0.8 \linewidth]{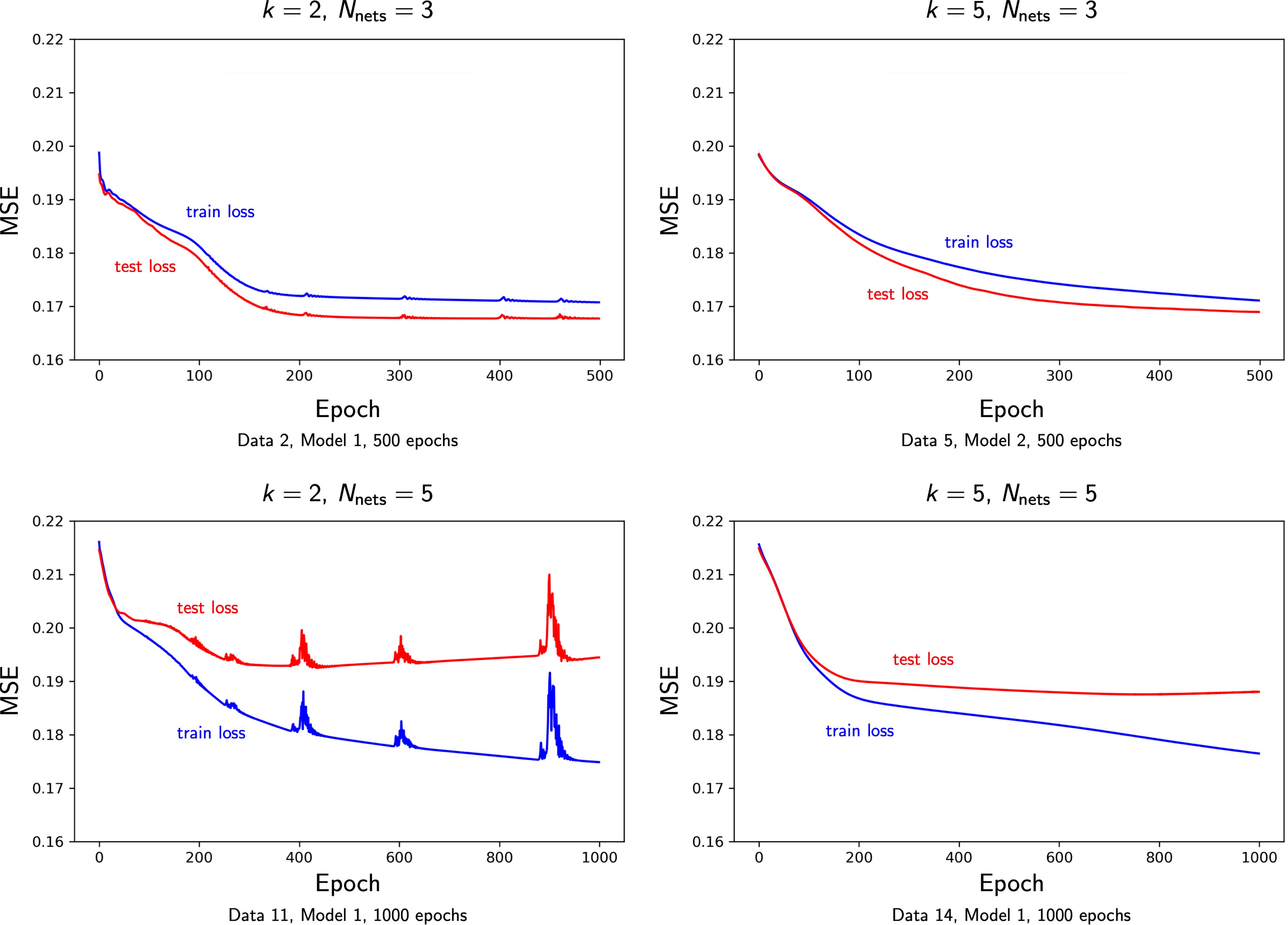}
}
\caption{
A selection of loss curves during training for a some proposed deep models with training hyperparameters.
\label{fig:floss}
}
\end{center}
\end{figure}

Training accuracy is calculated based on the performance of our deep learning models within each data group in the generated database. 
For a given dataset we have $N=2500$ data groups that form a set of different routing problems in the same routing environment. 
As a result, each data group is associated with a unique 1-layer routing solution $S^1$.
The actual optimal net ordering $no$ of a given $S^1$ corresponding to a data group is obtained using the simplified criteria summarized in Section \sref{sec:score}.
The accuracy of our deep learning models depends on whether the actual optimal net ordering $no$ for $S^1$ corresponding to a given data group is matched by the predicted optimal net ordering $\hat{no}$ for $S^1$ obtained from one of our deep learning models.
Accordingly, the overall accuracy of the deep learning model can be calculated following, 
\beal{es30a01}
\text{accuracy}(\%) = 
\frac{n_{\text{match}}}{N} 
~~(\times 100 \%)
~,~
\eea 
where $N$ is the number of data groups defined on the same routing problem parameters discussed in Section \sref{sec:41}, and $n_{\text{match}}$ is the number of data groups for which the machine learning predicted the correct optimal net ordering $\hat{no} = no$.
\tref{tab:t07} illustrates an example calculation of the model accuracy for $N=5$ data groups.
\fref{fig:floss} shows a selection of loss curves during the training of models 1 and 2 with selected training hyperparameters. 
\\

\begin{table}[ht!]
\begin{center}
\begin{tabular}{|c|c|c|c|c|}
\hline
data group & predicted order $\hat{no}$ & actual order $no$ & match & accuracy \\
\hline\hline
1 & $(1,2,3)$ & $(1,2,3)$ & $\circ$ & \\
2 & $(3,1,2)$ & $(3,1,2)$ & $\circ$ & \\
3 & $(1,3,2)$ & $(1,2,3)$ & $\times$ & \\
4 & $(2,1,3)$ & $(3,2,1)$ & $\times$ & \\
5 & $(1,3,2)$ & $(1,3,2)$ & $\circ$ & \multirow{-5}*{$60\%$}\\
\hline
\end{tabular}
\caption{
An example for calculating the accuracy for $N=5$ data groups corresponding to the same dataset. 
\label{tab:t07}
}
\end{center}
\end{table}

\section{Results \label{sec:5}}

\begin{figure}[ht!]
\begin{center}
\resizebox{0.6\hsize}{!}{
\includegraphics[height=8cm]{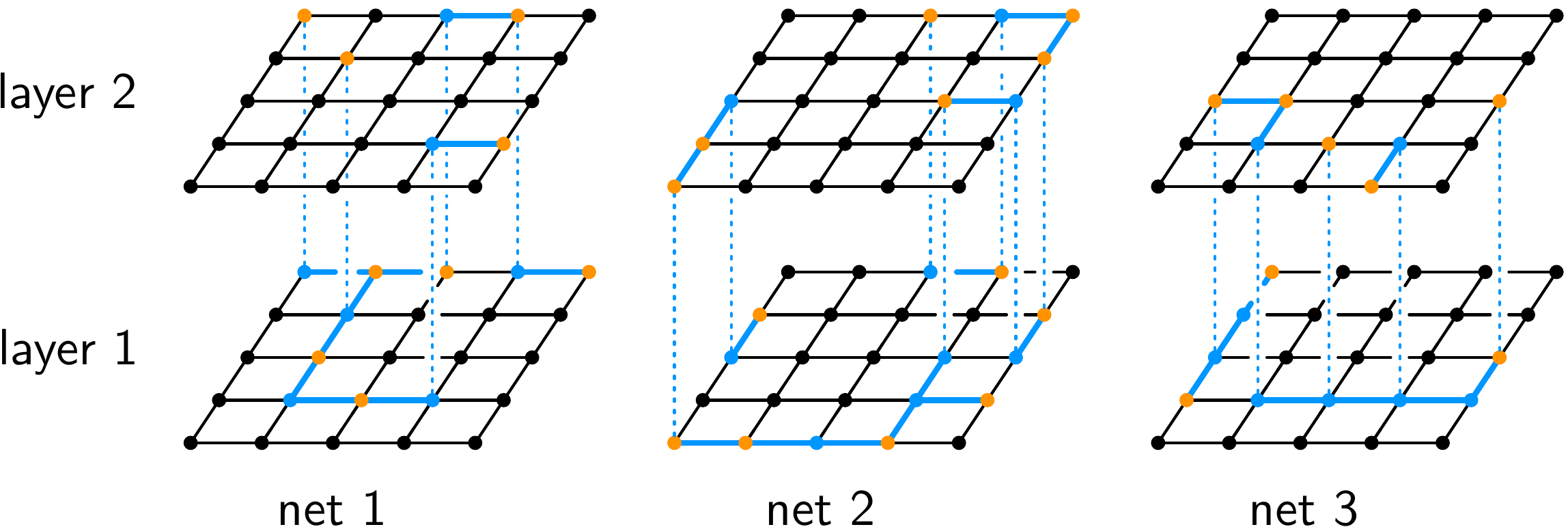} 
}
\caption{
Routing result $S^k$ for a $k=2$ layer routing environment with $N_{\text{nets}}=3$. 
The vertices $v_m^k$ that corresponding to pins are highlighted in orange and are restricted to a constant number of 15 per layer. 
The 1-layer routing solution $S^1$ was obtained through the Steiner tree algorithm. 
\label{fig:fsample1}}
 \end{center}
 \end{figure}

\begin{figure}[ht!]
\begin{center}
\resizebox{0.6\hsize}{!}{
\includegraphics[height=8cm]{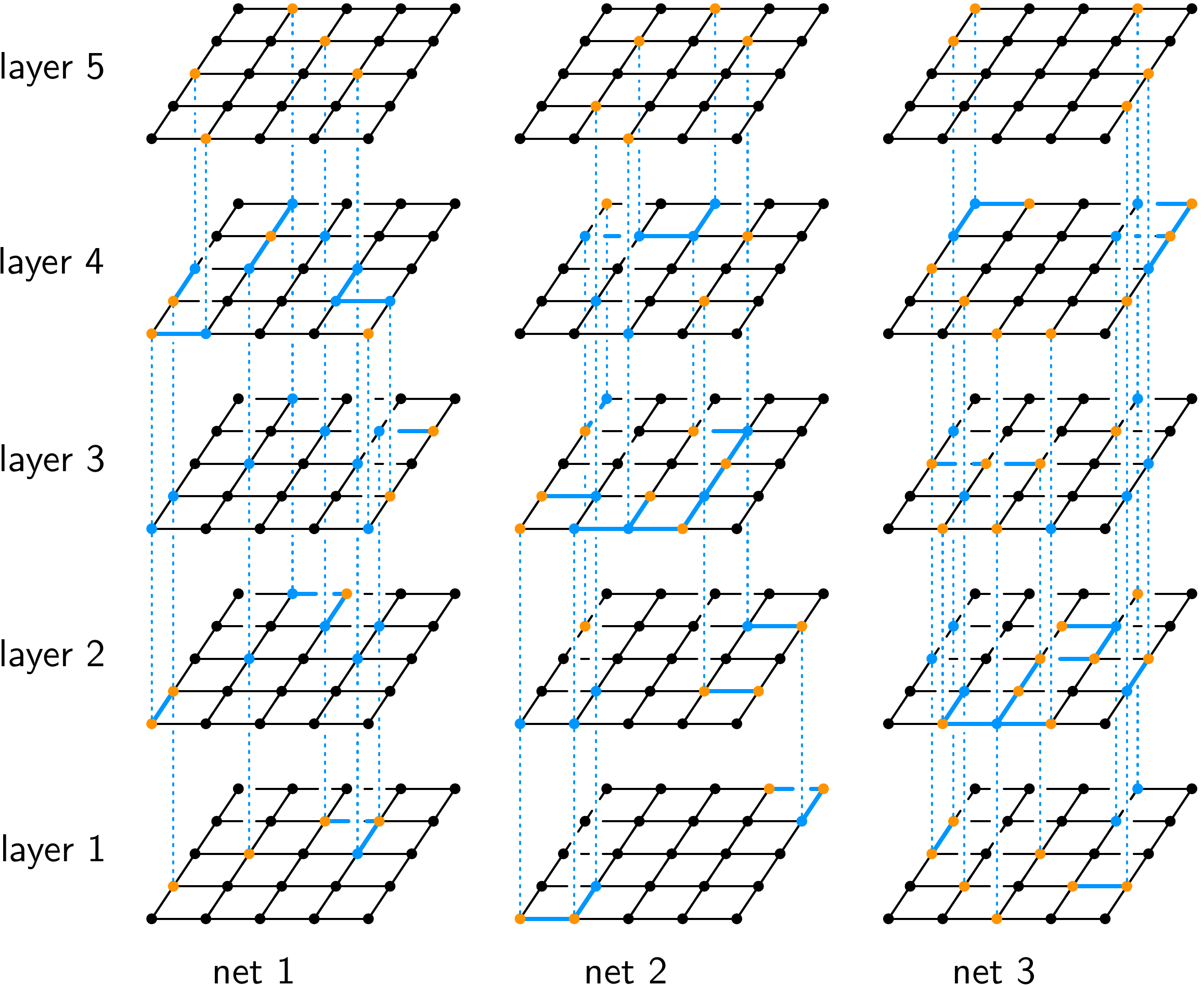} 
}
\caption{
Routing result $S^k$ for a $k=5$ layer routing environment with $N_{\text{nets}}=3$. 
The vertices $v_m^k$ that corresponding to pins are highlighted in orange and are restricted to a constant number of 15 per layer. 
The 1-layer routing solution $S^1$ was obtained through the Steiner tree algorithm. 
\label{fig:fsample2}}
 \end{center}
 \end{figure}

First, our experiments on our trained deep learning models suggest that our deep learning models perform better in predicting the optimal net ordering based on the optimality criteria summarized in Section \sref{sec:score} than 
the heuristic score function in \cite{lee2008congestion} or even a random choice of net ordering.
\fref{fig:fsample1} and \fref{fig:fsample2} show a selection of a 2-layer and a 5-layer routing result respectively using our proposed deep learning models for net ordering. 
\tref{tab:t08} summarizes the average accuracies of the predicted optimal net orderings for the 3 proposed deep learning models, the heuristic method based on \cite{lee2008congestion} and a random generator of net orderings.
Note that the random generator independent of the 1-layer router type and layer number $k$ generates a random net order $no$ which is only dependent on the number of nets $N_{\text{nets}}$. 
Furthermore, the heuristic method only uses the reduced set of features and therefore the average accuracies for the heuristic method do not distinguish between datasets with reduced or full set of features. 
The average accuracies are calculated over the $N=2500$ data groups for each of the 16 datasets. 
\tref{tab:t08} shows the highest average accuracies for a given $k$-layer environment and $N_{\text{nets}}$ number of nets in bold. 

For all datasets, our proposed deep learning models show much higher accuracy than the heuristic score function from \cite{lee2008congestion} or the random generator. 
When the number of nets is $N_{\text{nets}}=3$, the performance of our deep learning models is more than twice as the heuristic method or the random generator.
Furthermore, when the number of nets is $N_\text{nets}=5$, the performance of our deep learning models is more than 10 times higher than the heuristic method, which with higher net numbers performs nearly as bad as a simple random generator. 
Overall, as expected, smaller numbers of nets and layers decreases the complexity of the routing problem, resulting in an overall increase in the accuracy of the predicted optimal net ordering by our deep learning models.

\begin{table}[ht!]
\centering
\begin{tabular}{|c|ccc|c|c|}
\hline
 & \multicolumn{5}{c|}{accuracy($\%$)} \\
\hline
dataset & model 1 & model 2 & model 3 & heuristic & random\\
\hline\hline
Data 1 & 40.69 & \textbf{41.30} & 36.63 & & \\
Data 2 &  \textbf{41.30} & 40.89 & 33.82 & \multirow{-2}*{17.25} & \\ \cline{1-5}
Data 3 & 39.72 & 39.11 & 36.90 & & \\
Data 4 & 40.93 & 40.93 & 38.10 & \multirow{-2}*{15.44} & \\ \cline{1-5}
Data 5 & 40.44 & \textbf{42.05} & 34.21 & & \\
Data 6 & 40.04 & 41.25 & 33.20 & \multirow{-2}*{14.41} & \\ \cline{1-5}
Data 7 & 36.87 & 37.27 & 32.46 & & \\
Data 8 & 36.47 & 38.08 & 31.46 & \multirow{-2}*{14.51} & \multirow{-8}*{16.67}\\ \hline
Data 9 & 7.22 & \textbf{9.69} & 4.12 & & \\
Data 10 & 7.63 & 7.84 & 3.30 & \multirow{-2}*{0.70} & \\ \cline{1-5}
Data 11 & 8.81 & 7.79 & 2.66 & & \\
Data 12 & 8.40 & 8.20 & 2.87 & \multirow{-2}*{0.82} & \\ \cline{1-5}
Data 13 & 7.58 & 7.99 & 4.11 & & \\
Data 14 & 7.38 & 7.58 & 2.87 & \multirow{-2}*{0.25} & \\ \cline{1-5}
Data 15 & 7.76 & \textbf{8.16} & 4.29 &  & \\
Data 16 & 7.35 & 7.55 & 4.08 & \multirow{-2}*{0.78} & \multirow{-8}*{0.83}\\ \hline
\end{tabular}
\caption{
Summary of accuracy measurements for the proposed 3 deep learning models, the heuristic score function method based on previous work and a simple random generator. The accuracy measurements were performed over 16 datasets that each contain $N=2500$ randomly generated data groups associated to routing problems set under the same routing and environmental parameters specific to the dataset. 
\label{tab:t08}
}
\end{table}

Between the 3 different deep learning models that we propose in this work, models 1 and 2 perform better on average across all 16 datasets than model 3. 
This may indicate that the use of MSE as the loss function is better than the use of cross-entropy as suggested in Section \sref{sec:model}.
Moreover, according to the results in \tref{tab:t08}, it appears that the 3 deep learning models on average performed better across all the 16 datasets when the complete feature vector $\vec{f}(S^k)$ was used rather than the reduced feature vector $\vec{f}_{\text{reduced}}(S^k)$.
This indicates that the more we know about the 1-layer routing result $S^1$, the better our deep learning models perform in identifying the most optimal net ordering $no_h$ that results in the most optimal $k$-layer routing solution $S^k_h$.
We also note that overall the results based on the Kruskal algorithm (KA) for the $1$-layer router perform better than the results based on the Steiner tree (ST) algorithm. 

\begin{figure}[ht!]
\begin{center}
\resizebox{0.6\hsize}{!}{
\includegraphics[trim={0cm 0cm 0cm 0cm}, width=0.8 \linewidth]{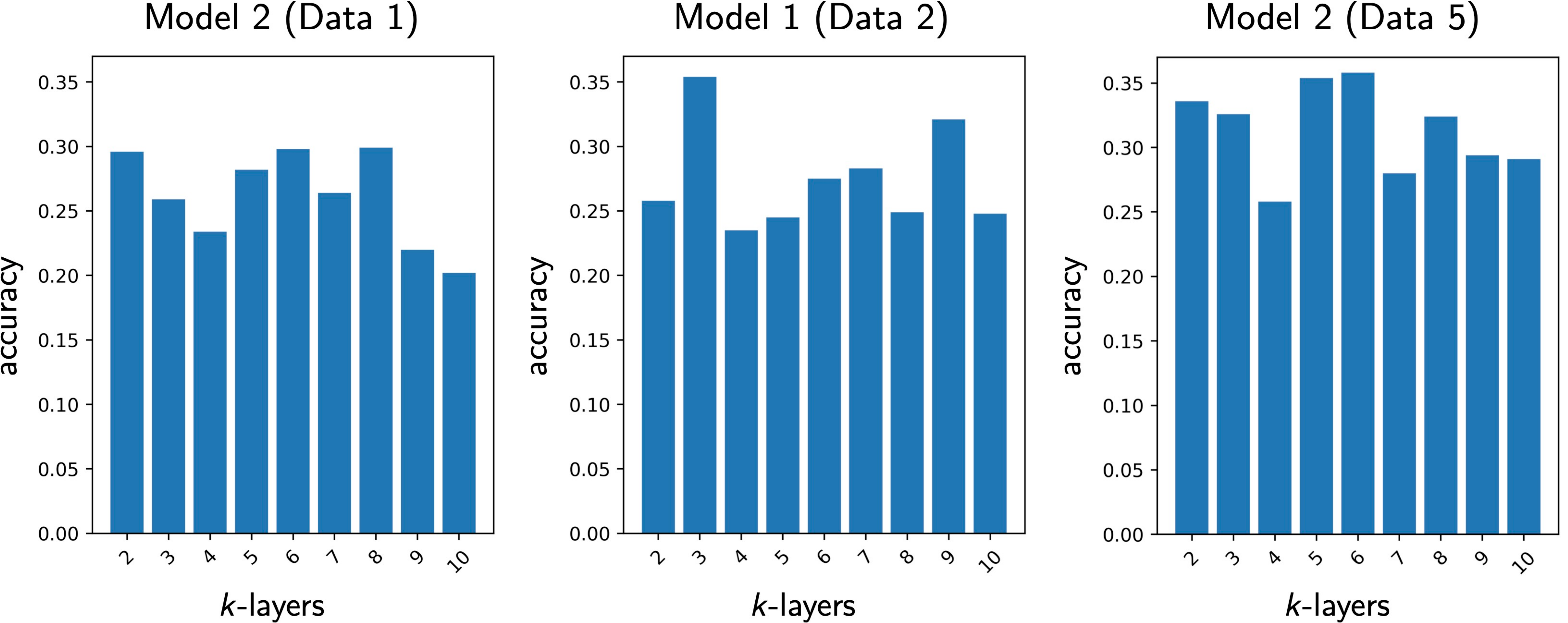}
}
\caption{
Models trained on a single dataset perform well for routing environments with different number of layers. 
\label{fig:ftransfer}
}
\end{center}
\end{figure}

Because our deep learning models significantly performed better than the previous heuristic score function method, we tested whether our trained deep learning models are transferable to other routing environments and problems.
In order to test transferability \cite{torrey2010transfer,zhuang2020comprehensive}, 
we select the best performing deep learning models in \tref{tab:t08} according to datasets that have $3$ nets.
Deep learning models 1 and 2 are trained using respectively datasets Data 2 and Data 1 in \tref{tab:t05} with $2$ layers and $3$ nets, and 
model 2 is trained with dataset Data 5 with $5$ layers and $3$ nets as defined in \tref{tab:t05}.
These deep learning models performed based for these specific datasets according to the results in \tref{tab:t08}.
We then used these 3 trained models and calculated their average accuracy on datasets with the same number of nets, but layer numbers ranging from 2 to 10. 
\fref{fig:ftransfer} summarizes the resulting average accuracy values for the 3 deep learning models, and we can see that there is no significant performance drop with increasing number of layers in the test datasets. 
This is even the case for model 2 trained on dataset Data 5, which with $5$ layers and $3$ nets is highly complex.
As a result, our proposed deep learning models appear to be highly transferable to routing environments with the same number of nets, but increasingly larger number of layers. 
This underlines the significant performance increase by our deep learning models and we hope to report on further improvements based on our deep learning models in the near future.
\\

\section{Conclusions \label{sec:6}}

In this work, we have proposed a new net ordering method based on machine learning techniques. 
Our new method outperforms previous heuristics-based methods proposed in \cite{lee2008congestion} based on the experiments conducted as part of this work.
Moreover, our experiments have also shown that indeed net ordering significantly affects the overall layer assignment process of the 2-dimensional routing result in global routing.
As a result, our machine learning based method significantly improved the global routing algorithm first proposed in \cite{lee2008congestion} for semiconductor chip and package design.

As discussed in section \sref{sec:2}, 
net ordering in global routing with layer compression plays a crucial role in the global routing process in semiconductor package design. 
We however emphasize that our work does not optimize other optimizeable steps in global routing with layer compression, including the algorithm for $2d$ routing after layer compression or the discretization of the routing environment. 
Our work concentrates on optimizing the net ordering part of the global routing process with layer compression.
A small improvement as we achieved in this work in the sequence of optimizeable steps in the global routing process has a cumulative positive effect for the overall global routing result. 

To give an example, we expect to be able to further optimize the global routing process by applying machine learning techniques to the actual layer assignment process described in \cite{lee2008congestion}. 
 like net ordering, the algorithm in \cite{lee2008congestion} for layer assignment of the 2-dimensional routing results is based on a heuristics-based minimum via cost function that measures the cost of lifting a certain segment of the 2-dimensional routing result to a $k$-th layer by assessing the cost of adding a via connection from layer 1 to layer $k$. 
This heuristics-based minimum cost function indeed can be replaced by a machine learning model that optimizes the minimum cost much more effectively than a hard-coded cost function based on heuristics. 
As a result, we believe that our improvements achieved in this work can be combined with further optimized routing steps such as the $2d$ routing step or the routing environment discretization step. We hope to explore these improvements in future work.

Additionally, we expect that further optimization can be achieved in net ordering if the number of layers and nets increases beyond the scope of the training data that we have used as part of this work. 
Furthermore, we hope to further improve our machine learning model by testing it against other model architectures such as convolutional neural networks (CNN) and graph neural networks (GNN). 
We hope to explore these directions in future work. 

As a final comment, we would like to emphasize that semiconductor package design 
heavily relies on physical design constraints.
These may include the efficiency of the design in dispersing heat generated by the components of the semiconductor package
and
the minimum distance between connections in order to avoid signal interference.
Evaluating the overall performance of a particular routing design under the collection of these physical constraints in itself is a challenging problem, which we plan to explore in future work.

\section*{Data Availabillity}
The data generated or analyzed during this study are available from the corresponding author on reasonable request.

\section*{Acknowledgements}

The authors were supported by a UNIST Industry Research Project (2.220916.01) funded by Samsung SDS in Korea,
as well as a UNIST UBSI Research Fund (1.220123.01, 1.230065.01) and the BK21 Program (``Next Generation Education Program for Mathematical Sciences'', 4299990414089) funded by the Ministry of Education in Korea and the National Research Foundation of Korea (NRF).
C.-H. L. is supported by a National Research Foundation of Korea (NRF) grant funded by the Korean government (MSIT) (NRF-2022R1F1A1064487).
R.-K. S. is supported by a Basic Research Grant of the National Research Foundation of Korea (NRF-2022R1F1A1073128).
He is also supported by a Start-up Research Grant for new faculty at UNIST (1.210139.01) and a UNIST AI Incubator Grant (1.230038.01).
\\
\\
This version of the
article has been accepted for publication, after peer review but is not the Version of
Record and does not reflect post-acceptance improvements, or any corrections. The Version of Record is
available online at: https://doi.org/10.1038/s41598-024-82226-9 .


\bibliography{mybib}      
             

\end{document}

%% file: pref.tex
\newcommand{\be}{\begin{equation}}
\newcommand{\ee}{\end{equation}}
\newcommand{\beq}{\begin{equation}}
\newcommand{\beql}[1]{\begin{equation}\label{#1}}
\newcommand{\eeq}{\end{equation}}
\newcommand{\ba}{\begin{array}}
\newcommand{\ea}{\end{array}}
\newcommand{\bea}{\begin{eqnarray}}
\newcommand{\beal}[1]{\begin{eqnarray}\label{#1}}
\newcommand{\eea}{\end{eqnarray}}
\newcommand{\ben}{\begin{enumerate}}
\newcommand{\een}{\end{enumerate}}
\newcommand{\bean}{\begin{eqnarray*}}
\newcommand{\eean}{\end{eqnarray*}}
\newcommand{\eref}[1]{(\ref{#1})}
\newcommand{\sref}[1]{\S\ref{#1}}
\newcommand{\tref}[1]{Table~\ref{#1}}

\newcommand{\fref}[1]{Figure \ref{#1}}
\newcommand{\btab}[1]{\begin{tabular}{#1}}
\newcommand{\etab}{\end{tabular}}